%% file: main.tex
\documentclass[10pt,twocolumn,letterpaper]{article}

\usepackage[pagenumbers]{cvpr} 

\input{preamble}
\definecolor{cvprblue}{rgb}{0.21,0.49,0.74}
\usepackage[pagebackref,breaklinks,colorlinks,allcolors=cvprblue,urlcolor=magenta]{hyperref}
\usepackage{multirow} 
\usepackage{listings}
\usepackage{xcolor}
\usepackage{graphicx}
\usepackage{pifont}
\usepackage{amssymb}
\usepackage{lscape}

\def\paperID{****} 
\def\confName{CVPR}
\def\confYear{2026}

\title{ActAvatar: Temporally-Aware Precise Action Control for Talking Avatars}

\author{
Ziqiao Peng$^{1*}$\quad Yi Chen$^{2}$\quad Yifeng Ma$^2$\quad Guozhen Zhang$^2$\quad Zhiyao Sun$^2$\\ 
Zixiang Zhou$^{2}$\quad Youliang Zhang$^{2}$\quad Zhengguang Zhou$^2$\quad Zhaoxin Fan$^1$ \\
Hongyan Liu$^{3\ddagger}$ \quad Yuan Zhou$^{2\dagger}$\quad Qinglin Lu$^{2\ddagger}$\quad Jun He$^{1\ddagger}$ \\
$^1$Renmin University of China\quad $^2$Tencent Hunyuan\quad  $^3$Tsinghua University \\
\url{https://ziqiaopeng.github.io/ActAvatar/}
}

\begin{document}

\twocolumn[{
\maketitle
\begin{center}
    \captionsetup{type=figure}
    \vspace{-18pt}
    \includegraphics[width=0.97\textwidth]{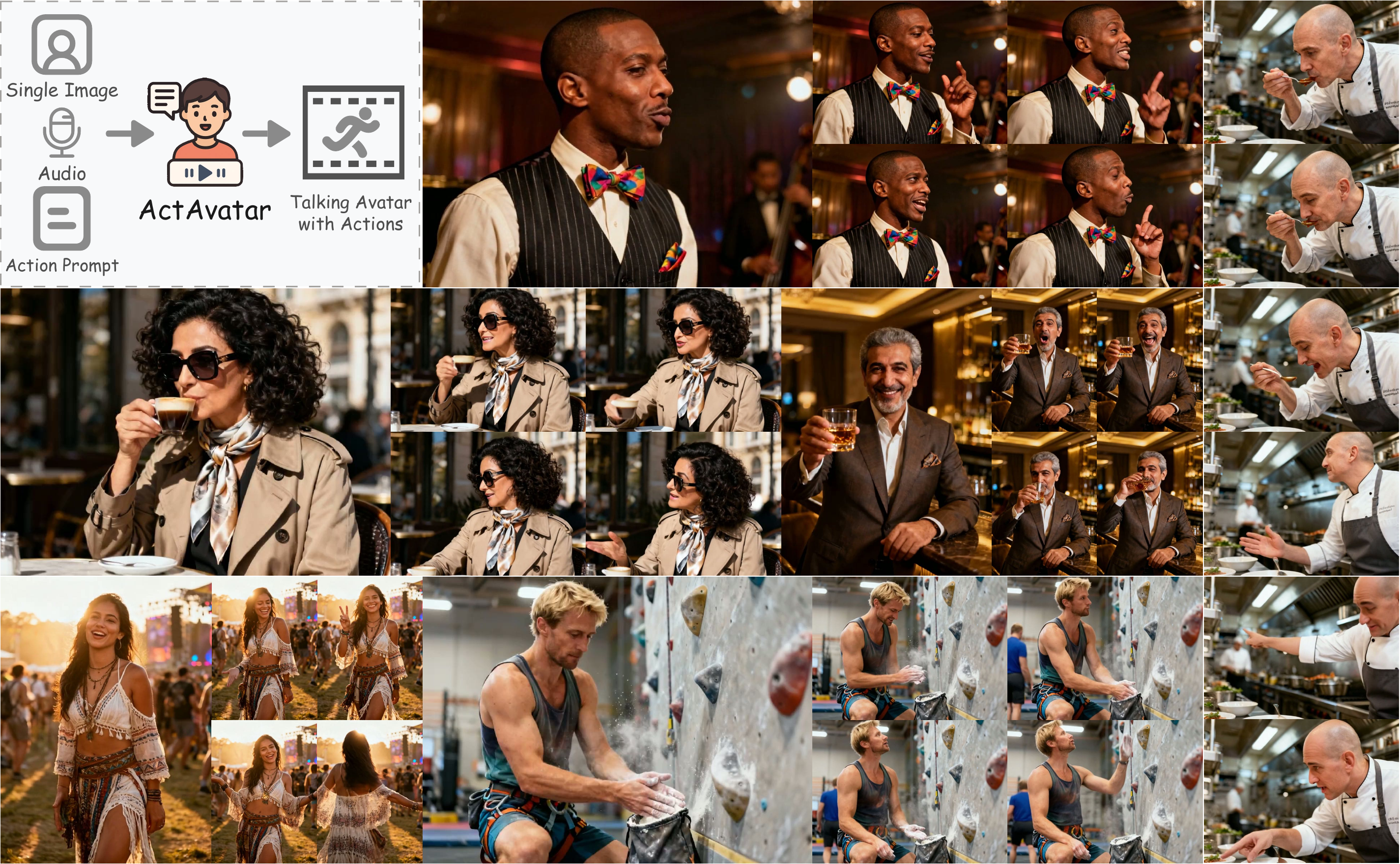}
    \vspace{-0.6em}
    \caption{ActAvatar generates talking avatars with precise, temporally-aligned actions across diverse scenarios and identities. Through structured text prompts, our method controls what actions to perform and when to perform them, while maintaining accurate lip synchronization with the audio.}
    \label{fig:1}

\end{center}
}]

\renewcommand{\thefootnote}{\fnsymbol{footnote}}
\footnotetext[1]{Work done during an internship at Tencent Hunyuan.}
\footnotetext[2]{Project Leader.}
\footnotetext[3]{Corresponding Author.}

\maketitle

\input{sec/0_abstract}    
\input{sec/1_intro}

\input{sec/2_related}
\input{sec/3_method}
\input{sec/4_experiment}

\input{sec/5_conclusion}


{
    \small
    \bibliographystyle{ieeenat_fullname}
    \bibliography{main}
}

\end{document}

%% file: sec/0_abstract.tex
\begin{abstract}

\vspace{-8pt}

Despite significant advances in talking avatar generation, existing methods face critical challenges: insufficient text-following capability for diverse actions, lack of temporal alignment between actions and audio content, and dependency on additional control signals such as pose skeletons. We present ActAvatar, a framework that achieves phase-level precision in action control through textual guidance by capturing both action semantics and temporal context. Our approach introduces three core innovations: (1) Phase-Aware Cross-Attention (PACA), which decomposes prompts into a global base block and temporally-anchored phase blocks, enabling the model to concentrate on phase-relevant tokens for precise temporal-semantic alignment; (2) Progressive Audio-Visual Alignment, which aligns modality influence with the hierarchical feature learning process—early layers prioritize text for establishing action structure while deeper layers emphasize audio for refining lip movements, preventing modality interference; (3) A two-stage training strategy that first establishes robust audio-visual correspondence on diverse data, then injects action control through fine-tuning on structured annotations, maintaining both audio-visual alignment and the model's text-following capabilities. Extensive experiments demonstrate that ActAvatar significantly outperforms state-of-the-art methods in both action control and visual quality.

\end{abstract}

%% file: sec/1_intro.tex
\section{Introduction}
\label{sec:intro}

Talking avatar generation has become increasingly important for applications ranging from virtual assistants~\cite{peng2023emotalk} and digital entertainment to online education~\cite{ye2023geneface} and telepresence systems~\cite{peng2025synctalk++}. Recent advances in diffusion models~\cite{wan2025wan,kong2024hunyuanvideo} have significantly improved the visual quality of generated talking avatars~\cite{cui2024hallo3,wei2025mocha,wang2025fantasytalking,peng2025omnisync, lin2025omnihuman,gan2025omniavatar,zhang2025uniavgen}. 

However, existing methods~\cite{zhang2023sadtalker, tu2025stableavatar, peng2024synctalk, lin2025omnihuman} face three critical limitations. First, while current models can generate plausible hand movements~\cite{zhou2025exges}, they struggle to accurately execute specific actions described in prompts due to treating the entire prompt uniformly~\cite{kong2025let}, where action-related descriptions compete with scene descriptions for attention. Second, actions may appear at arbitrary moments rather than synchronizing with semantically relevant speech segments~\cite{zhang2025uniavgen}. This temporal drift arises because standard conditioning mechanisms lack explicit temporal structure, causing attention to diffuse uniformly across time. Third, many methods resort to explicit control modalities such as pose skeleton sequences~\cite{meng2025echomimicv2, cui2025hallo4}, which increase pipeline complexity and limit the ability to generate novel actions.

These limitations reveal a fundamental challenge: establishing explicit correspondences between language semantics (what actions to perform), temporal windows (when to perform them), and audio cues (how they relate to speech). Moreover, text-driven action generation and audio-driven lip synchronization represent competing objectives that can interfere during generation. When both modalities exert strong influence simultaneously, the model struggles to balance conflicting signals, often resulting in degraded action quality or compromised lip-sync accuracy~\cite{wang2020makes}. In addition, fine-tuning pre-trained models on domain-specific data to improve audio-visual alignment often leads to catastrophic forgetting, where the model's original text-following capabilities are weakened or lost entirely. 

To address these challenges, we present \textbf{ActAvatar}, a framework that achieves temporally-aware, precise action control for talking avatar generation through textual guidance. Our key insight is threefold: (1) structured prompt organization with temporal anchors enables learned phase-conditioned attention dynamics for temporal-semantic alignment; (2) progressive modality influence prevents interference between text-driven action generation and audio-driven lip synchronization; (3) staged training preserves multiple capabilities by decoupling.

Our approach introduces three synergistic technical innovations. We propose \textbf{Phase-Aware Cross-Attention (PACA)}, which decomposes prompts into hierarchically structured phases with explicit temporal grounding. By organizing textual descriptions into a global base block and phase-specific blocks with temporal anchors, PACA enables the model to concentrate attention on temporally-relevant tokens during corresponding time windows. 

Second, we develop \textbf{Progressive Audio-Visual Alignment}, which addresses the interference between text-driven action generation and audio-driven lip synchronization by aligning modality influence with the hierarchical feature learning process. Early transformer layers prioritize text conditioning to establish overall action structure. As generation progresses to deeper layers, audio emphasis gradually increases, allowing refinement of lip movements after the primary action framework has been determined. This progressive strategy prevents modality interference while ensuring both accurate action generation and precise lip sync.

Third, we propose a \textbf{two-stage training strategy} that addresses the capability preservation challenge through task decomposition. We first establish robust audio-visual alignment in Stage 1, then introduce temporal action control in Stage 2. This staged approach enables the model to learn action control as a compositional extension of existing capabilities rather than through destructive parameter updates.

\begin{figure*}
\vspace{-1em}
\begin{center}
   \includegraphics[width=1.\linewidth]{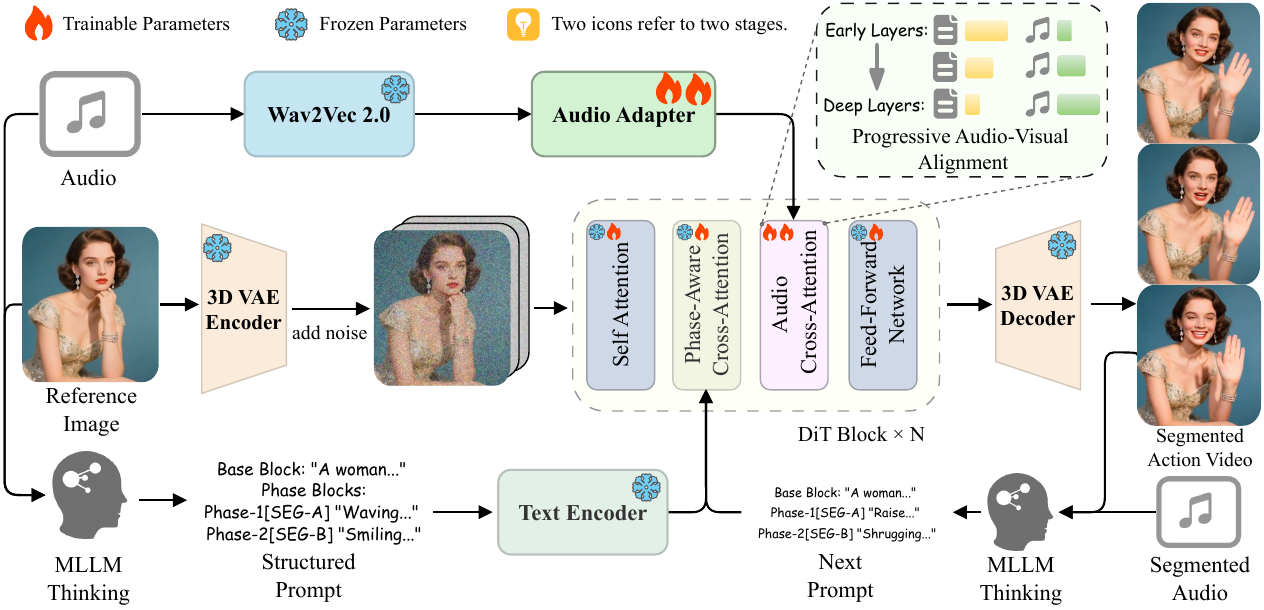}
\end{center}
\vspace{-1em}
    \caption{\textbf{Overview of ActAvatar.} Given an audio input and reference image, ActAvatar generates temporally-controlled action videos guided by structured prompts automatically generated from MLLM.}
\label{fig:pipe}
\vspace{-0.5em}
\end{figure*}

In summary, our contributions are as follows:

\begin{itemize}
    \item \textbf{Phase-Aware Cross-Attention:} A hierarchical prompt decomposition mechanism with temporal anchors that enables learned phase-conditioned attention dynamics for precise temporal-semantic alignment.
    
    \item \textbf{Progressive Audio-Visual Alignment:} A depth-aware modality scaling mechanism that aligns text and audio influence with hierarchical feature learning, preventing interference between text-driven action generation and audio-driven lip synchronization.
        
    \item \textbf{Staged Capability Preservation Training:} A two-stage training paradigm that addresses catastrophic forgetting through task decomposition, enabling compositional capability extension while preserving both audio-visual alignment and text-following capabilities.
        
\end{itemize}

%% file: sec/2_related.tex
\section{Related Work}
\label{sec:related}

\subsection{Video Generation Models}

Text-to-video generation~\cite{wu2023tune,singer2022make,rao2024modelgrow,lin2024open,zhang2025arbitrary,peng2024controlnext} has witnessed remarkable progress with the advent of diffusion models. Early works such as Video Diffusion Models~\cite{ho2022video} and Imagen Video~\cite{ho2022imagen} established foundational frameworks by extending image diffusion to the temporal domain through 3D U-Net architectures and factorized spatiotemporal attention. Subsequent methods like AnimateDiff~\cite{guo2023animatediff} and Stable Video Diffusion~\cite{blattmann2023stable} achieved substantial improvements in visual quality and motion coherence by leveraging pre-trained text-to-image models and introducing specialized temporal modeling modules. 

Recently, video generation models have shifted towards Transformer-based~\cite{vaswani2017attention} architectures to achieve better scalability. HunyuanVideo~\cite{kong2024hunyuanvideo} creates a comprehensive training framework that advances efficient model training and inference, Wan~\cite{wan2025wan} builds a complete and open video generation model suite that drives the development of the open-source community, and SkyReels v2~\cite{chen2025skyreels} further extends video generation duration to infinite-length generation.

\subsection{Talking Avatar Generation}

Talking avatar generation~\cite{cui2024hallo3,wang2024styletalk++,peng2025dualtalk,ma2025talkclip, meng2025echomimicv3, yang2025flowerdance,yang2025megadance,ma2023dreamtalk,peng2023selftalk,zhang2025morpheus, chen2025hunyuanvideo,zhou2024meta, hu2025ggtalker,tu2025stableavatar,wu2024vgg,ma2023styletalk,wang2024v, ma2025playmate2} aims to synthesize realistic human video with synchronized lip movements and natural expressions driven by audio input. EMO~\cite{tian2024emo} pioneered audio-conditioned video diffusion for talking heads, introducing audio cross-attention for lip synchronization. EchoMimic~\cite{chen2025echomimic}further improved controllability through multi-modal conditioning combining audio with visual references. Methods like Echomimic V2~\cite{meng2025echomimicv2} and Hallo 4~\cite{cui2025hallo4} resort to explicit pose guidance through skeleton sequences, introducing additional annotation requirements and limiting the naturalness of language-based interaction. MultiTalk~\cite{kong2025let} and HunyuanVideo-Avatar~\cite{chen2025hunyuanvideo} extend these approaches to natural motion generation, but suffer from poor text-following capability. 

Recent works further optimize this task, such as Kling-Avatar~\cite{ding2025kling} and OmniHuman 1.5~\cite{jiang2025omnihuman}. These methods rely on strong video generation models to achieve relatively good action generation capabilities, but still perform diffusion on global prompts, failing to achieve precise temporal action control. AgentAvatar~\cite{wang2023agentavatar} try to introduce a timeline method, but only generate facial expressions.
Our method addresses this limitation by introducing structured temporal prompts that enable phase-level precision in action control through hierarchical prompt decomposition and phase-aware attention mechanisms.

%% file: sec/3_method.tex
\section{Method}

\subsection{Overview}

ActAvatar aims to achieve precise temporal action control in talking avatar generation through structured prompt conditioning. Given an audio sequence $\mathbf{a} \in \mathbb{R}^{T_a \times D_a}$ and a reference image $\mathbf{I}_{\text{ref}} \in \mathbb{R}^{H \times W \times 3}$, our goal is to generate a video $\mathbf{V} = \{\mathbf{I}_t\}_{t=1}^{T}$ where the avatar exhibits accurate lip synchronization with the audio and performs specific actions at semantically appropriate temporal windows as described in the structured prompt $\mathbf{P}$. The structured prompt is automatically generated by a Multimodal Large Language Model (MLLM, e.g., Qwen3-Omni~\cite{xu2025qwen3})  based on the input image and audio content.

Our framework builds upon an image-to-video diffusion backbone and introduces three synergistic components: (1) Phase-Aware Cross-Attention (PACA), which enables temporal-semantic alignment through hierarchical prompt decomposition and learned phase-conditioned attention dynamics; (2) Progressive Audio-Visual Alignment, which prevents modality interference by aligning text and audio influence with the hierarchical feature learning process; and (3) Two-Stage Training preserves both audio-visual correspondence and text-following capabilities through task decomposition. Figure~\ref{fig:pipe} illustrates the overall architecture.

\subsection{Phase-Aware Cross-Attention}

\subsubsection{Hierarchical Prompt Decomposition}

Standard talking avatar methods condition generation on a single global prompt $\mathbf{P}_{\text{global}}$ that describes the overall scene. This flat representation lacks temporal structure, causing semantic diffusion where action-related information is uniformly distributed across all timesteps. To address this, we introduce hierarchical decomposition that explicitly encodes temporal grounding:

\begin{equation}
\mathbf{P} = \{\mathbf{P}_{\text{base}}, \{\mathbf{P}_k, \mathcal{T}_k\}_{k=1}^{K}\},
\end{equation}
where $\mathbf{P}_{\text{base}}$ is a global base block encoding time-invariant scene semantics including identity descriptors, environmental context, affective state, stylistic constraints, and global motion characteristics. Each phase block $\mathbf{P}_k$ describes temporally-localized actions within a designated temporal window $\mathcal{T}_k = [\tau_k^{\text{start}}, \tau_k^{\text{end}}]$ specified in normalized time coordinates. The base block establishes stable scene context, while phase blocks introduce temporal specificity through explicit time anchoring.

For example: \textit{Base}: ``A woman in business attire speaking professionally"; \textit{Phase-1 [0-2s]}: ``Gestures outward with open palm"; \textit{Phase-2 [2-4s]}: ``Points downward to emphasize detail."

\subsubsection{Phase-Conditioned Attention Dynamics}

Let $\mathbf{x}_f \in \mathbb{R}^{N \times D}$ denote the video latent features at frame index $f$, where $N$ is the number of spatial tokens and $D$ is the feature dimension. The structured prompt $\mathbf{P}$ is first encoded into a token sequence through a pre-trained text encoder (umT5-XXL), yielding $\mathbf{C} = \{\mathbf{c}_i\}_{i=1}^{M} \in \mathbb{R}^{M \times D_c}$.

\noindent\textbf{Phase Position Encoding.} To explicitly encode phase membership and enhance phase-conditioned attention, we introduce learnable phase position embeddings. For each token $\mathbf{c}_i$ belonging to phase $k$, we add a phase-specific positional bias:

\begin{equation}
\mathbf{c}_i' = \mathbf{c}_i + \mathbf{e}_k,
\end{equation}
where $\mathbf{e}_k \in \mathbb{R}^{D_c}$ is a learnable phase embedding for phase $k$. The embeddings are zero-initialized to ensure identity behavior at the start of training.
These phase embeddings provide an inductive bias that encourages the model to distinguish between base and phase-specific tokens in the attention mechanism.

In standard cross-attention, queries $\mathbf{Q}_f = \mathbf{x}_f \mathbf{W}_Q$, keys $\mathbf{K} = \mathbf{C}' \mathbf{W}_K$, and values $\mathbf{V} = \mathbf{C}' \mathbf{W}_V$ are computed from the phase-augmented token embeddings:

\begin{equation}
\text{Attention}(\mathbf{Q}_f, \mathbf{K}, \mathbf{V}) = \text{softmax}\left(\frac{\mathbf{Q}_f \mathbf{K}^T}{\sqrt{D}}\right) \mathbf{V}.
\end{equation}

Through training on temporally-annotated data, the model learns to concentrate attention on phase-relevant tokens when frame $f$ (normalized to video time $\tau \in [0, T_{\text{video}}]$) falls within the corresponding temporal range $\mathcal{T}_k$, achieving temporal-semantic alignment through the combination of temporal anchors and phase position embeddings.

\subsection{Progressive Audio-Visual Alignment}

Text-driven action generation and audio-driven lip synchronization represent competing objectives that can interfere during the generation process. To address this, we introduce progressive audio-visual alignment that aligns modality influence with the hierarchical feature learning process in diffusion transformers.

Diffusion transformers naturally follow a coarse-to-fine feature learning hierarchy: early layers capture global structure and layout, while deeper layers refine local details and high-frequency features. We leverage this characteristic to prevent modality interference by progressively scaling audio influence across transformer blocks.

For transformer block $\ell \in \{1, \ldots, L\}$, we apply depth-aware scaling to the audio cross-attention residual:

\begin{equation}
\mathbf{x}_\ell \leftarrow \mathbf{x}_\ell + f(\ell) \cdot \mathbf{r}_{\text{audio}}^\ell,
\end{equation}
where the scaling function is:

\begin{equation}
f(\ell) = \left(\frac{\ell}{L}\right)^\gamma,
\end{equation}
with $\gamma > 1$. This creates progressive amplification of audio influence in deeper layers.

The design aligns modality influence with the generation hierarchy. Early layers ($\ell \ll L$, small $f(\ell)$) prioritize text conditioning to establish overall action structure—body pose, hand trajectory, and gesture type. During this phase, audio influence remains minimal, allowing text to dominate the generation of action semantics without interference from audio signals. As generation progresses to deeper layers ($\ell \to L$, larger $f(\ell)$), audio emphasis gradually increases, enabling precise refinement of lip movements and facial articulation after the primary action framework has been determined.

This progressive strategy prevents modality interference by ensuring that text and audio operate in complementary rather than competing regimes: text establishes the action structure in early layers where coarse features dominate, while audio refines lip articulation in deep layers where high-frequency details emerge. 

\subsection{Two-Stage Training Strategy}

Our two-stage training strategy addresses the poor text-following capability observed in previous talking avatar methods by decoupling audio-visual learning from temporal action control. This staged approach enables the model to first establish robust audio-visual correspondence, then integrate precise temporal action semantics without compromising either capability.

\subsubsection{Stage 1: Audio Adapter Training and Extraction}

Stage 1 establishes robust audio-visual correspondence by training on diverse and large-scale talking-head videos. We adopt the Flow Matching training paradigm~\cite{lipman2022flow}. Given the original latent representation $\mathbf{x}_0$ (data) and pure noise $\mathbf{x}_1 \sim \mathcal{N}(0, \mathbf{I})$, we construct the optimal transport flow path:

\begin{equation}
\mathbf{x}_t = (1 - t) \mathbf{x}_0 + t \mathbf{x}_1,
\end{equation}
where $t \in [0, 1]$ is the flow time uniformly sampled from the training timestep sequence. The model is trained to predict the velocity field:

\begin{equation}
\mathbf{v}_{\text{target}} = \mathbf{x}_1 - \mathbf{x}_0,
\end{equation}
which represents the direction from data to noise. The complete Stage 1 loss is:

\begin{equation}
\mathcal{L}_{\text{stage1}} = \mathbb{E}_{\mathbf{x}_0 , t, \mathbf{x}_1} \left[ \left\| \mathbf{v}_\theta(\mathbf{x}_t, t, \mathbf{C}_{\text{brief}}, \mathbf{A}) - (\mathbf{x}_1 - \mathbf{x}_0) \right\|^2 \right],
\end{equation}
where $\mathbf{C}_{\text{brief}}$ represents brief text captions (e.g., ``A woman speaking"), and $\mathbf{A}$ are audio embeddings from Wav2Vec 2.0~\cite{baevski2020wav2vec}. The audio adapter consists of an audio projection module (mapping audio encodings to frame-aligned tokens) and audio cross-attention layers (fusing audio semantics via frame-wise attention).

Crucially, we freeze the base text-to-video backbone parameters $\theta_{\text{base}}$ and only train the audio adapter parameters $\theta_{\text{audio}}$:

\begin{equation}
\theta_{\text{stage1}} = \{\theta_{\text{audio}}\}, \quad \theta_{\text{base}} \text{ frozen}.
\end{equation}

This selective training preserves the pre-learned text-to-image correspondence and spatial attention patterns while integrating audio conditioning. The diverse data distribution ensures that audio-visual alignment generalizes across different speakers, emotions, languages, and environmental conditions. After training, we extract the learned audio cross-attention modules as a pretrained audio adapter for Stage 2.

\subsubsection{Stage 2: Temporally-Aware Action Control}

Stage 2 injects temporal action control through structured annotations. We construct a dataset via: (1) applying DWPose~\cite{yang2023effective} to compute motion magnitude and selecting videos with significant movement; (2) using a multimodal large language model to generate hierarchical prompts with base blocks and phase-specific descriptions with temporal anchors. This yields a focused dataset with high-quality structured annotations.

We construct the Stage 2 model by starting with a base image-to-video model and injecting the pretrained audio adapter from Stage 1. The training objective maintains the Flow Matching form:

\begin{equation}
\mathcal{L}_{\text{stage2}} = \mathbb{E}_{\mathbf{x}_0, t, \mathbf{x}_1} \left[ \left\| \mathbf{v}_\theta(\mathbf{x}_t, t, \mathbf{C}_{\text{PACA}}, \mathbf{A}) - (\mathbf{x}_1 - \mathbf{x}_0) \right\|^2 \right],
\end{equation}
where $\mathbf{C}_{\text{PACA}}$ is the hierarchically-structured prompt encoding with phase position embeddings.

In Stage 2, we adopt full fine-tuning that simultaneously optimizes both speech synchronization and action control:

\begin{equation}
\theta_{\text{stage2}} = \{\theta_{\text{base}}, \theta_{\text{audio}}, \theta_{\text{PACA}}\}.
\end{equation}

The two-stage approach preserves both audio-visual correspondence and text-following capabilities: Stage 1 establishes robust lip synchronization on diverse data with frozen backbone, while Stage 2's full fine-tuning on structured annotations enables precise temporal action control.

\begin{table*}[t]
\centering
\caption{Quantitative comparison on HDTF Test Set. Best results in \textbf{bold}, second best \underline{underlined}.}
\label{tab:hdtf_results}
\resizebox{0.94\textwidth}{!}{
\begin{tabular}{l|cccc|cc|ccc}
\toprule
\multirow{2}{*}{Method} & \multicolumn{4}{c|}{Visual Quality} & \multicolumn{2}{c|}{Lip Sync} & \multicolumn{3}{c}{Model Info} \\
& FID$\downarrow$ & FVD$\downarrow$ & IQA$\uparrow$ & ASE$\uparrow$ & Sync-C$\uparrow$ & Sync-D$\downarrow$ & Params & Resolution & Time\\
\midrule
Hallo3~\cite{cui2024hallo3} & 30.424 & 318.451 & 3.883 & 2.488 & 6.994 & 8.647 & 5B & 480p & 32 min \\
FantasyTalking~\cite{wang2025fantasytalking} & 25.615 & 458.733 & 3.955 & 2.663 & 4.179 & 11.067 & 14B & 720p & 83 min\\
EchoMimic v3~\cite{meng2025echomimicv3} & 43.544 & 595.805 & 4.027 & 2.690 & 4.555 & 10.804 & 1.3B & 480p & 7 min\\
HunyuanVideo-Avatar~\cite{chen2025hunyuanvideo} & 24.515 & 322.844 & 4.054 & \underline{2.693} & 7.647 & \underline{7.564} & 13B & 720p & 74 min\\
MultiTalk~\cite{kong2025let} & 25.578 & 368.719 & 3.981 & 2.570 & 7.610 & 7.645 & 14B & 480p & 45 min\\
OmniAvatar~\cite{gan2025omniavatar} & 24.398 & 374.594 & 4.088 & 2.664 & \textbf{7.986} & 7.696 & 14B & 480p & 36 min\\
StableAvatar~\cite{tu2025stableavatar} & \underline{23.742} & 323.601 & 3.984 & 2.572 & 5.457 & 10.151 & 1.3B & 480p & 12 min\\
Wan-S2V~\cite{gao2025wan} & 23.850 & \textbf{299.491} & \underline{4.108} & 2.684 & 7.462 & 7.745 & 14B & 720p & 68 min\\
\midrule
ActAvatar (Ours) & \textbf{23.471} & \underline{301.064} & \textbf{4.120} & \textbf{2.714} & \underline{7.663} & \textbf{7.545} & 5B & 720p & 16 min\\
\bottomrule
\end{tabular}}
\end{table*}

\begin{table*}[t]
\centering
\caption{Quantitative comparison on Action Bench. Best results in \textbf{bold}, second best \underline{underlined}.}
\label{tab:action_results}
\resizebox{\textwidth}{!}{
\begin{tabular}{l|cc|cc|ccccc}
\toprule
\multirow{3}{*}{Method} & \multicolumn{2}{c|}{Lip Sync} & \multicolumn{2}{c|}{Visual Quality} & \multicolumn{5}{c}{Gemini-Based Action Metrics} \\
& Sync-C$\uparrow$ & Sync-D$\downarrow$ & IQA$\uparrow$ & ASE$\uparrow$ & \begin{tabular}[c]{@{}c@{}}Hit@ \\ Segment\end{tabular}$\uparrow$ &  \begin{tabular}[c]{@{}c@{}}Action \\ Accuracy\end{tabular}$\uparrow$ & \begin{tabular}[c]{@{}c@{}}Temporal \\ Correctness\end{tabular}$\uparrow$ & \begin{tabular}[c]{@{}c@{}}Action \\ Quality\end{tabular}$\uparrow$ & \begin{tabular}[c]{@{}c@{}}Hand \\ Clarity\end{tabular}$\uparrow$ \\
\midrule
Hallo3~\cite{cui2024hallo3} & 4.288 & 10.266 & 4.462 & 3.531 & 0.756 & 5.299 & 6.541 & 6.793 & 7.844 \\
FantasyTalking~\cite{wang2025fantasytalking} & 4.209 & 10.948 & 4.510 & 3.560 & 0.728 & 4.995 & 6.435 & 6.501 & 8.147 \\
EchoMimic v3~\cite{meng2025echomimicv3} & 3.199 & 11.896 & 4.301 & 3.543 & 0.764 & 4.717 & 6.684 & 5.817 & 6.115 \\
HunyuanVideo-Avatar~\cite{chen2025hunyuanvideo} & 6.251 & 8.862 & 4.759 & \underline{3.725} & 0.674 & 3.977 & 5.491 & 6.609 & 8.044 \\
MultiTalk~\cite{kong2025let} & 6.654 & \underline{8.356} & 4.446 & 3.477 & 0.574 & 3.490 & 4.654 & 5.455 & 7.584 \\
OmniAvatar~\cite{gan2025omniavatar} & \underline{6.765} & 8.785 & 4.569 & 3.610 & \underline{0.818} & \underline{5.505} & \underline{7.032} & \underline{7.147} & 8.042 \\
StableAvatar~\cite{tu2025stableavatar} & 4.035 & 11.615 & \underline{4.760} & 3.658 & 0.758 & 4.802 & 6.576 & 6.722 & 7.768 \\
Wan-S2V~\cite{gao2025wan} & 6.473 & 8.401 & 4.701 & 3.647 & 0.754 & 4.934 & 6.465 & 6.630 & \underline{8.168} \\
\midrule
ActAvatar (Ours) & \textbf{6.893} & \textbf{8.246} & \textbf{4.814} & \textbf{3.743} & \textbf{0.854} & \textbf{5.971} & \textbf{7.353} & \textbf{7.671} & \textbf{8.483} \\
\bottomrule
\end{tabular}}
\end{table*}

%% file: sec/4_experiment.tex
\section{Experiments}

\subsection{Experimental Setup}

\subsubsection{Implementation Details} We implement ActAvatar using PyTorch on 40 NVIDIA H20 GPUs. The backbone is Wan2.2-TI2V-5B with 30 DiT blocks. The audio encoder is Wav2Vec 2.0 and text encoder is umT5-XXL. Stage 1 trains for 20K steps and Stage 2 for 14K steps, both with batch size 40, learning rate $5 \times 10^{-6}$, and AdamW optimizer. For Progressive Audio-Visual Alignment, we set $\gamma=1.5$ in $f(\ell) = (\ell/30)^\gamma$. We generate 125-frame videos (5s at 25 FPS) at 704$\times$1280 resolution using flow-matching sampling with 40 steps and classifier-free guidance scale 5.0 for both text and audio.

\begin{figure*}[t]
\centering
\includegraphics[width=\textwidth]{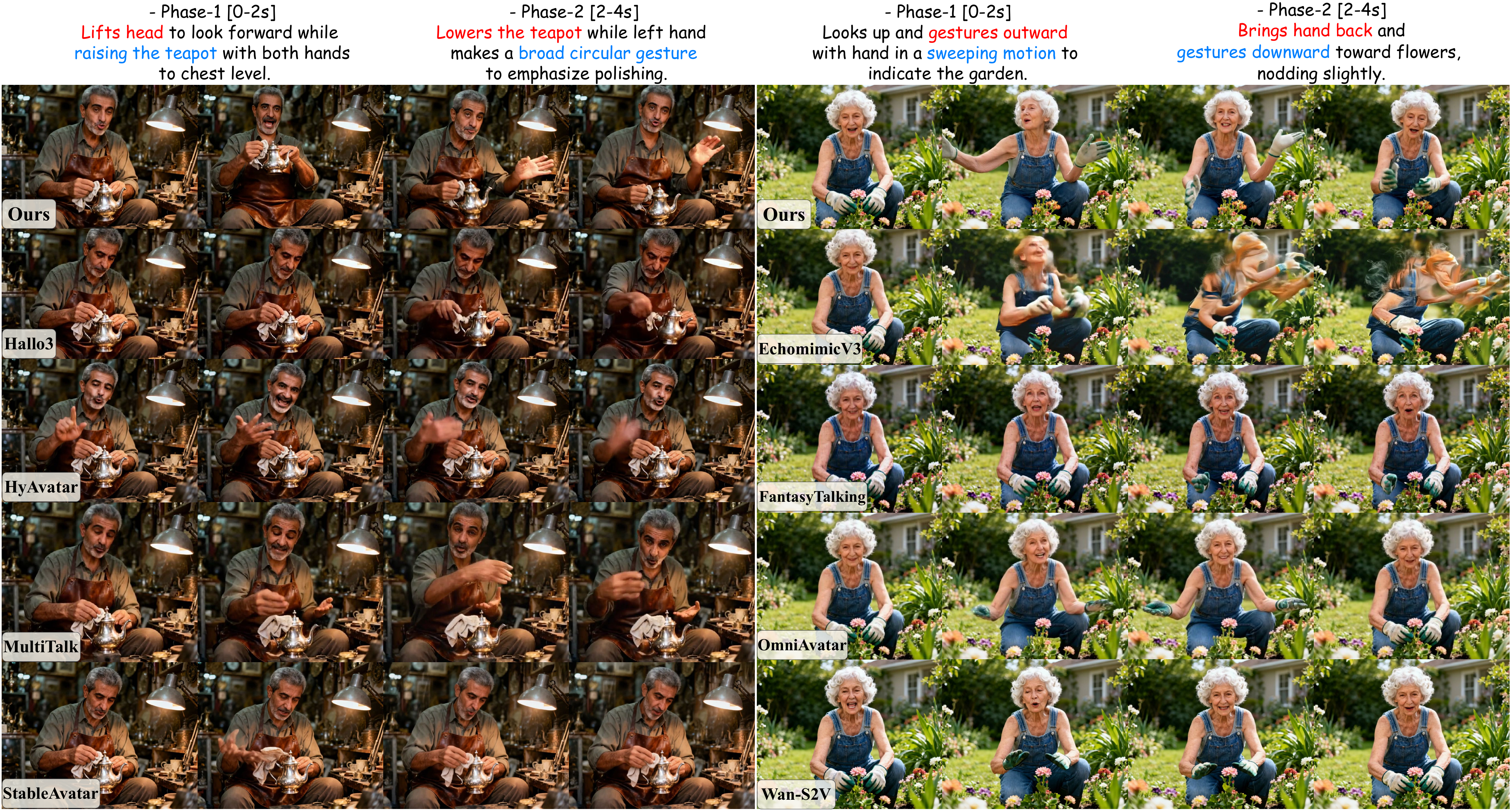}
\caption{Qualitative comparison with state-of-the-art methods. ActAvatar accurately executes phase-specific actions with correct timing and clear hand articulation, while competing methods show temporal misalignment, vague motions, or degraded hand quality.}
\label{fig:comparison}
\end{figure*}

\subsubsection{Datasets}

\noindent\textbf{Training Data.} For Stage 1 audio-driven lip synchronization training, we utilize 500K diverse talking head videos from OpenHumanVid~\cite{li2025openhumanvid} and SpeakerVid~\cite{zhang2025speakervid}, covering varied speakers, emotions, and speaking styles. For Stage 2, we construct a structured annotation dataset through DWPose-based motion selection and MLLM-based prompt generation, yielding 100K samples with phase-level temporal annotations. The detailed construction pipeline is provided in the supplementary material.

\noindent\textbf{Evaluation Data.} We evaluate on two test sets: (1) \textit{HDTF Test Set}~\cite{zhang2021flow}, containing 100 high-quality talking-head videos (5s each) focusing on lip synchronization quality, as it includes only upper body without hand movements; (2) \textit{Action Bench}, our constructed benchmark consisting of 200 samples with diverse action instructions. Each sample includes a reference image, TTS-synthesized speech, and structured prompts with action annotations.The prompts are MLLM-generated with human verification. 

\subsubsection{Evaluation Metrics}

We evaluate ActAvatar using comprehensive metrics. For lip sync, we use Sync-C and Sync-D from SyncNet~\cite{chung2016out} (higher Sync-C and lower Sync-D indicate better alignment). For visual quality, we report FID and FVD (lower is better), and use Q-Align~\cite{wu2023q} for video quality (IQA) and aesthetics (ASE) (higher is better). For action control, we develop a Gemini-based evaluation framework that provides: (1) Action Occurrence (AO): whether the described action appears; (2) Action Accuracy (AA): how well the action matches the description (0-10); (3) Temporal Correctness (TC): whether the action occurs in the specified time window (0-10); (4) Action Quality (AQ): overall naturalness of execution (0-10); (5) Hand Clarity (HC): hand quality (0-10). We report Hit@Segment (H@S) as the percentage of phases where AO=1, and mean scores for AA, TC, AQ, and HC. To ensure robustness, we run the Gemini evaluation 5 times for each video and report the average scores. The evaluation prompts and framework construction are provided in the supplementary material.

\subsection{Quantitative Evaluation}

We present comprehensive quantitative comparisons on both HDTF Test Set and Action Bench. Table~\ref{tab:hdtf_results} shows results on HDTF, focusing on lip synchronization and visual quality in natural talking scenarios. And table~\ref{tab:action_results} presents results on Action Bench, evaluating action control capabilities alongside lip-sync and visual quality.

\noindent\textbf{Performance on HDTF.} ActAvatar achieves the best visual quality (FID: 23.471, IQA: 4.120, ASE: 2.714) while maintaining competitive lip synchronization (Sync-C: 7.663, Sync-D: 7.545). Operating at 720p with only 5B parameters, ActAvatar matches or exceeds 14B models running at lower resolutions. The lip-sync performance is on par with the best methods, demonstrating that our two-stage training successfully preserves audio-visual alignment.

\noindent\textbf{Performance on Action Bench.} ActAvatar demonstrates substantial advantages in action control. We achieve the highest H@S (0.854), significantly outperforming methods with other methods (OmniAvatar: 0.818, EchoMimic v3: 0.764). Gemini-based metrics show consistent superiority: AA (5.971 vs. 5.505), TC (7.353 vs. 7.032), AQ (7.671 vs. 7.147), and HC (8.483 vs. 8.168). Remarkably, ActAvatar achieves the best lip synchronization on Action Bench (Sync-C: 6.893), demonstrating that PACA enables precise action control without sacrificing audio-visual alignment.

\noindent\textbf{Inference Efficiency.} We further evaluate inference speed on a single H20 GPU. ActAvatar generates 5-second videos in 16 minutes, achieving more than 4× speedup over comparable methods (Wan-S2V: 68 min, FantasyTalking: 83 min, HunyuanVideo-Avatar: 74 min) while maintaining superior quality. With 8× H20 GPUs, generation time reduces to just 2 minutes per 5-second video. Although lightweight models like EchoMimic v3 (7 min) and StableAvatar (12 min) are faster, they show significantly degraded quality and lip-sync. ActAvatar's 5B model achieves optimal quality-efficiency balance, delivering 14B-level performance at competitive speeds.

\subsection{Qualitative Analysis}

\begin{figure}[t]
\centering
\includegraphics[width=0.95\linewidth]{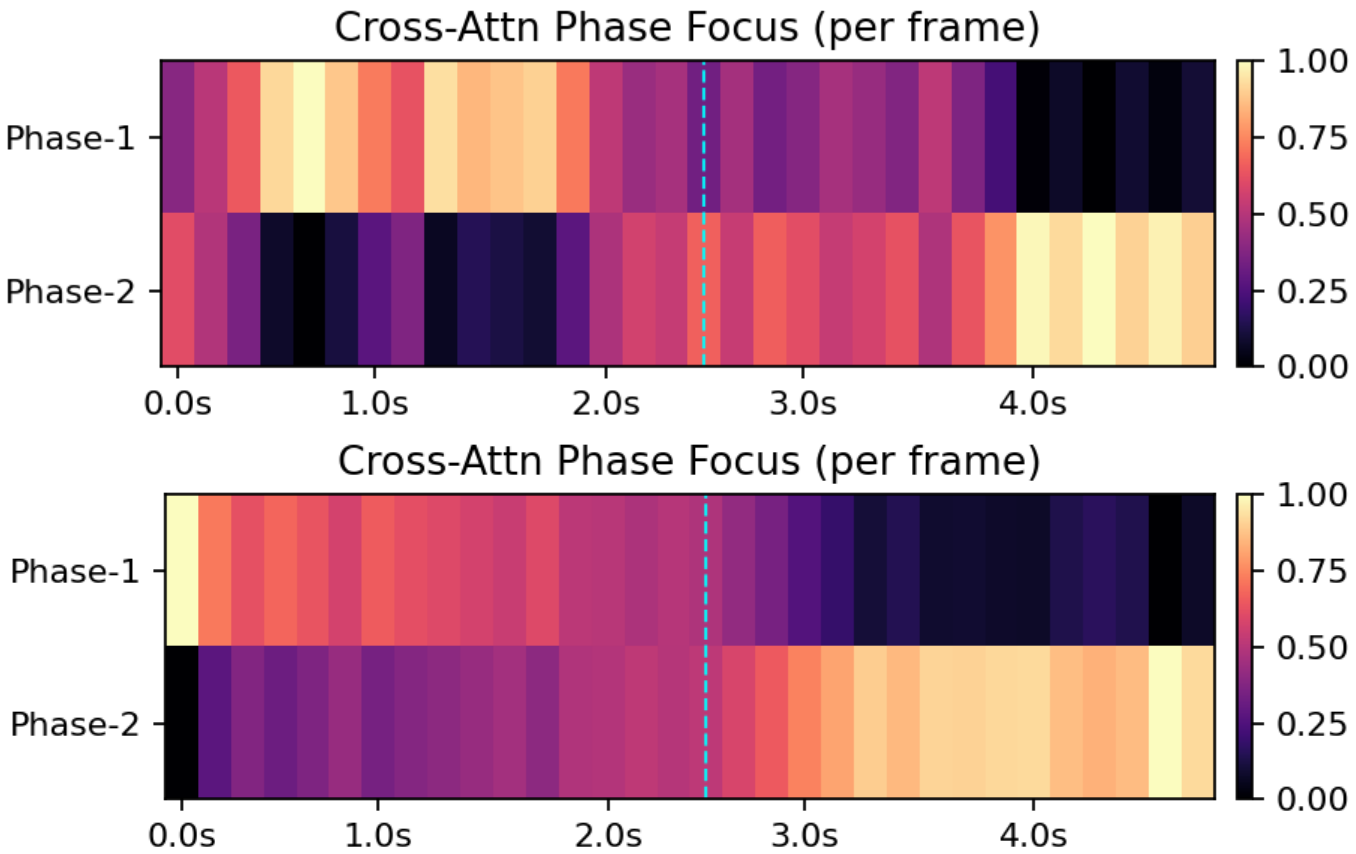}
\caption{Cross-attention phase focus at layer 5 (top) and layer 20 (bottom). Deeper layers show sharper phase separation.}
\label{fig:attention_vis}
\end{figure}

\noindent\textbf{Attention Visualization.} We visualizes the cross-attention distribution across time for different phase blocks at layer 5 (top) and layer 20 (bottom) (Figure~\ref{fig:attention_vis}). For a two-phase prompt, the attention naturally concentrates on Phase-1 tokens during the first half and shifts to Phase-2 tokens during the second half. This phase-conditioned attention pattern emerges, validating that our PACA module enables learned temporal-semantic correspondence.

Comparing layer 5 and layer 20 reveals hierarchical refinement: early layers show coarse phase awareness with some overlap around the boundary, while deeper layers exhibit sharp phase separation with concentrated attention. This aligns with the coarse-to-fine feature learning in transformers and explains the high Temporal Correctness scores in quantitative evaluation.

\noindent\textbf{Visual Quality Comparison.} Figure~\ref{fig:comparison} presents side-by-side qualitative comparisons with state-of-the-art methods on two representative examples from Action Bench, each featuring two distinct action phases with structured prompts. ActAvatar successfully executes both action phases with clear temporal separation and natural transitions. In contrast, competing methods exhibit various limitations. Echomimic V3 completely collapses when faced with such large-scale motions. Due to an over-reliance on the reference frame input, Hunyuanvideo-Avatar generates artifacts such as a third hand. Hallo3, StableAvatar, and FantasyTalking have very weak responsiveness to text, producing almost exclusively lip motion. While MultiTalk exhibits some hand movement, the motion does not follow the content of the prompt.

\subsection{User Study}

To complement quantitative evaluation, we conduct a user study with 45 participants. Participants evaluate videos generated by ActAvatar and competing methods for the same prompt, rating five dimensions on a 0-5 scale: (1) Action-Prompt Alignment (APA): how well actions match prompt descriptions; (2) Action Quality (AQ): naturalness and expressiveness of movements; (3) Hand Clarity (HC): clarity of hand gestures; (4) Lip Sync Accuracy (LSA): synchronization between lip movements and audio; (5) Overall Video Quality (OVQ): overall visual fidelity and realism. Each participant evaluates 30 videos with randomized order. Table~\ref{tab:user_study} presents the mean scores across all participants.

\begin{table}[t]
\centering
\caption{User study results. }
\label{tab:user_study}
\resizebox{\linewidth}{!}{
\begin{tabular}{l|ccccc}
\toprule
Method & APA $\uparrow$ & AQ $\uparrow$ & HC $\uparrow$ & LSA $\uparrow$ & OVQ $\uparrow$ \\
\midrule
EchoMimic v3~\cite{meng2025echomimicv3} & 3.43 & 3.25 & 3.21 & 2.45 & 2.68 \\
HunyuanVideo-Avatar~\cite{chen2025hunyuanvideo} & 2.84 & 3.12 & 3.68 & 3.67 & 3.45 \\
MultiTalk~\cite{kong2025let} & 3.19 & 3.16 & 3.95 & 3.41 & 3.58 \\
OmniAvatar~\cite{gan2025omniavatar} & 3.75 & 3.92 & 4.08 & 3.56 & 3.61 \\
Wan-S2V~\cite{gao2025wan} & 3.67 & 3.78 & 4.12 & 3.68 & 4.04 \\
\midrule
ActAvatar (Ours) & \textbf{4.03} & \textbf{4.15} & \textbf{4.22} & \textbf{3.89} & \textbf{4.18} \\
\bottomrule
\end{tabular}
}
\end{table}

ActAvatar achieves the highest scores across all dimensions, with particularly strong performance in action-related metrics. For Action-Prompt Alignment (APA: 4.03), ActAvatar substantially outperforms all baselines, confirming that PACA enables perceptually recognizable temporal action control. Hand Clarity (HC: 4.22) is ActAvatar's strongest dimension, validating that the model maintains clear hand articulation during dynamic gestures.

Notably, the user study rankings closely align with our Gemini-based quantitative metrics on Action Bench. This  correlation validates the reliability of our Gemini-based evaluation framework—the automated assessments align well with human perceptual judgments, demonstrating that Gemini can accurately capture fine-grained action quality and temporal alignment that matter to human evaluators.

\subsection{Ablation Studies}

\begin{figure}[t]
\centering
\includegraphics[width=\linewidth]{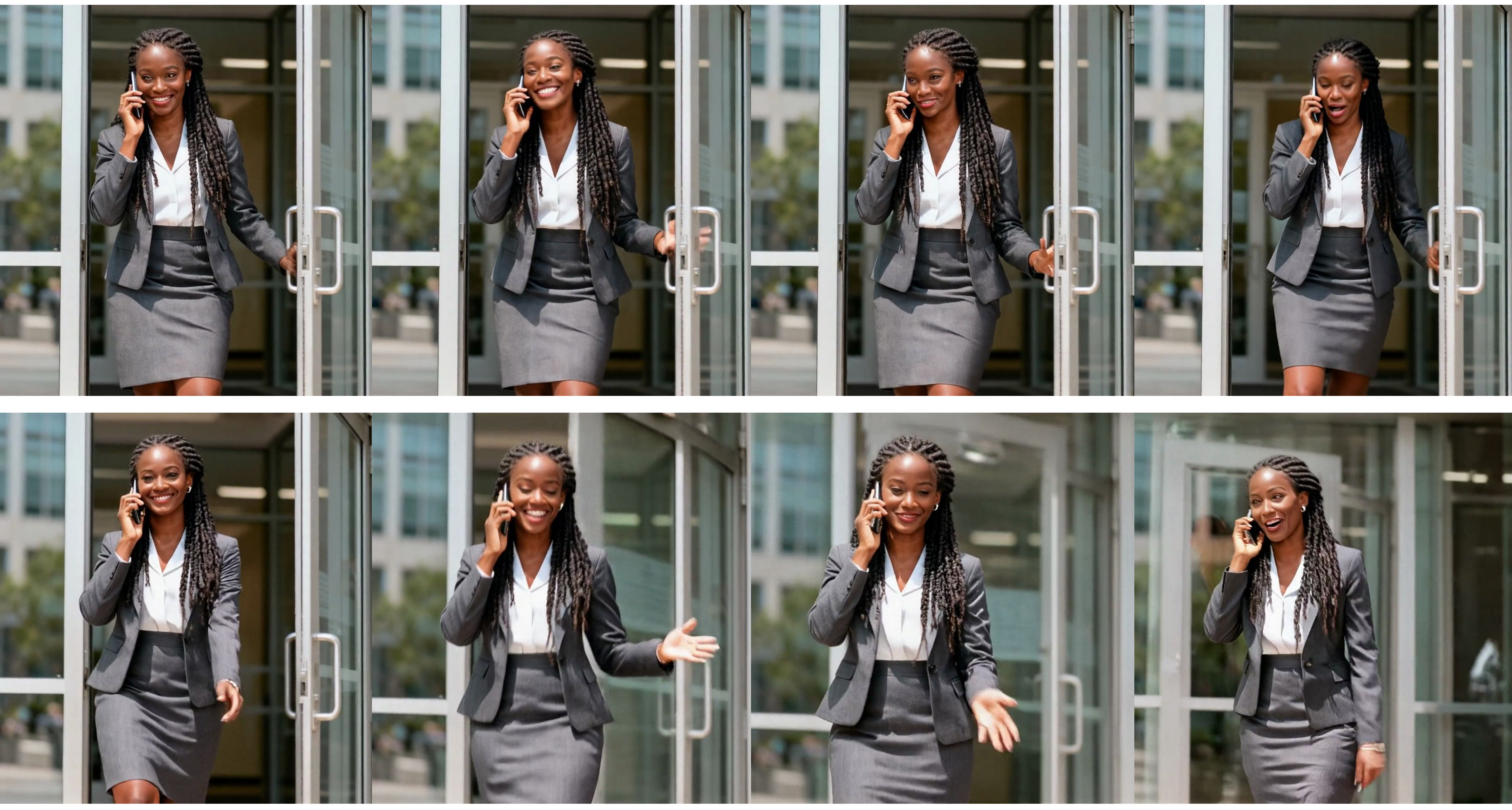}
\caption{Ablation study on PACA. \textbf{Top}: Without PACA, the avatar remains static throughout the sequence. \textbf{Bottom}: With PACA, the avatar naturally walks forward.}
\label{fig:ablation}
\end{figure}

To clarify the contributions of each core component in our framework, we conduct an ablation study targeting three key modules. Table~\ref{tab:ablation} validates the effectiveness of each core component in ActAvatar.

\begin{table}[t]
\centering
\caption{Ablation study of key components on Action Bench.}
\label{tab:ablation}
\resizebox{1\linewidth}{!}{
\begin{tabular}{l|c|ccccc}
\toprule
Configuration & Sync-C$\uparrow$ & H@S$\uparrow$ & AA$\uparrow$ & TC$\uparrow$ & AQ$\uparrow$ & HC$\uparrow$ \\
\midrule
Base (Global Prompt) & 6.37 & 0.725 & 3.91 & 6.47 & 6.21 & 7.68 \\
+ PACA & 6.39 & 0.829 & 5.78 & 7.12 & 7.48 & 8.36 \\
+ PACA + Progressive Alignment & 6.57 & 0.831 & 5.75 & 7.10 & 7.52 & 8.47 \\
+ Two-Stage Training (Full) & \textbf{6.89} & \textbf{0.854} & \textbf{5.97} & \textbf{7.35} & \textbf{7.67} & \textbf{8.48} \\
\bottomrule
\end{tabular}
}
\end{table}

\noindent\textbf{PACA.} Adding PACA to the base model substantially improves action control (H@S: 0.725 $\rightarrow$ 0.829), with corresponding gains in Action Accuracy (AA: 3.91 $\rightarrow$ 5.78) and Temporal Correctness (TC: 6.47 $\rightarrow$ 7.48). Figure~\ref{fig:ablation} provides visual compare: without PACA, the avatar remains static, while with PACA, dynamic motion emerges naturally. PACA effectively enables phase-conditioned attention for temporal-semantic alignment.

\noindent\textbf{Progressive Audio Alignment.} Adding depth-aware audio scaling improves lip synchronization (Sync-C: 6.39 $\rightarrow$ 6.57) while maintaining strong action performance (H@S: 0.831). The progressive strategy prevents modality interference by allowing text to dominate action generation in early layers while audio refines lip articulation in deeper layers.

\noindent\textbf{Two-Stage Training.} The complete two-stage training strategy provides further improvements across all metrics, achieving the best lip synchronization (Sync-C: 6.89) and action control (H@S: 0.854, AA: 5.97, TC: 7.35). This validates that decoupling audio-visual learning from action control injection is essential for maintaining both capabilities simultaneously.

%% file: sec/5_conclusion.tex
\section{Conclusion}

We present ActAvatar, a framework achieving precise temporal action control in talking avatar generation through textual guidance. By introducing Phase-Aware Cross-Attention, Progressive Audio-Visual Alignment, and a two-stage training strategy, ActAvatar addresses fundamental limitations of poor text-following and temporal misalignment in existing methods. Extensive experiments demonstrate that our approach significantly outperforms the SOTA methods in action accuracy, lip synchronization, and visual fidelity. Our work shows that structured textual conditioning can achieve phase-level temporal precision without additional control signals, opening new possibilities for controllable talking avatar generation.